\documentclass[11pt,a4paper]{article}
\usepackage[hyperref]{eacl2021}
\usepackage{times}
\usepackage{latexsym}
\usepackage{graphicx}

\usepackage{microtype}

\aclfinalcopy 

\usepackage[acronym,xindy,toc,numberedsection]{glossaries}
\newglossary[nlg]{notation}{not}{ntn}{Notation} 	
\makeglossaries
\newacronym{gdpr}{GDPR}{General Data Protection Regulation}
\newacronym{rgpd}{RGPD}{Règlement Général sur la Protection des Données}
\newacronym{rgpd2}{GDPR}{General Data Protection Regulation}
\newacronym{tal}{TAL}{Traitement Automatique des Langues}
\newacronym{nlp}{NLP}{Natural Language Processing}
\newacronym{ai}{AI}{Artificial Intelligence}
\newacronym{ertim2}{ERTIM}{Computer Text and Multilingualism Research Team}
\newacronym{ertim}{ERTIM}{Equipe de Recherche Texte Informatique et Multilinguisme}
\newacronym{inalco}{INALCO}{Institut National des Langues et des Civilisations Orientales}
\newacronym{inalco2}{INALCO}{National Institute of Oriental Languages and Civilizations}
\newacronym{llca}{LLNA}{Linguistiques et Littérature Africaine}
\newacronym{llca2}{LLNA}{Linguistics and African Literature}
\newacronym{onu}{ONU}{Organisation des Nations Unies}
\newacronym{onu2}{UNO}{United Nations Organization}
\newacronym{xnd}{XND}{XML Ntealan Dictionaries}
\newacronym{api}{API}{Application Programming Interface}
\newacronym{rest}{REST}{Representational State Transfert}
\newacronym{nlu}{NLU}{Natural Language Understanding}
\newacronym{ocr}{OCR}{Optical Character Recognition}
\newacronym{dtd}{DTD}{Document Type Declaration}
\newacronym{nsn}{NSN}{NTeALan Social Network}
\newacronym{nlr}{NLR}{NTeALan Language Research}
\newacronym{nais}{NAIS}{NTeALan AI Services}
\newacronym{xml}{XML}{Extensible Markup Language}
\newacronym{sgml}{SGML}{Standard Generalized Markup Language}
\newacronym{lmf}{LMF}{Language Markup Framework}
\newacronym{tei}{TEI}{Text Encoding Initiative}
\newacronym{gml}{GML}{Generalized Markup Language}
\newacronym{formex}{FORMEX}{Formalized Exchange of Electronic Publications}
\newacronym{cern}{CERN}{Conseil européen pour la recherche nucléaire}
\newacronym{inria}{INRIA}{Institut National de Recherche en Informatique et en Automatique}
\newacronym{html}{HTML}{Hypertext Markup Language}
\newacronym{w3c}{W3C}{World Wide Web}
\newacronym{css}{CSS}{Cascading Style Sheets}
\newacronym{alpac}{ALPAC}{Automatic Language Processing Advisory Committee}
\newacronym{ta}{TA}{Traduction Automatique}
\newacronym{xsd}{XSD}{XML Schema Definition}
\newacronym{relax_ng}{Relax NG}{Regular Language for XML Next Generation}
\newacronym{iso}{ISO}{International Organization for Standardization}
\newacronym{docbook}{DocBook}{Regular Language for XML Next Generation}
\newacronym{ead}{EAD}{Encoded Archival Description}
\newacronym{poo}{POO}{Programmation Orientée Objet}
\newacronym{tlf}{TLF}{Trésor de la Langue Française}
\newacronym{rad}{RAD}{Rapid Application Development}
\newacronym{dilaf}{DILAF}{Dictionnaire Informatisé en Langues Africaines}
\newacronym{ndlf}{NDLF}{Nouveau Dictionnaire de la Langue Française}
\newacronym{os}{OS}{Operating System}
\newacronym{tlfi}{TLFI}{Trésor de la Langue Française informatisé}
\newacronym{decfc}{DECFC}{Dictionnaire Explicatif et Combinatoire du Français}
\newacronym{ntealan}{NTeALan}{New Technologies for Africans Languages}
\newacronym{lms}{LMS}{Learning Management System}
\newacronym{apil}{APIL}{Association des Professionnels des Industries de la Langue}
\newacronym{til}{TIL}{Traitement Informatique de la Langue}
\newacronym{ia}{IA}{Intélligence Artificielle}
\newacronym{saas}{SAAS}{Software as a Service}
\newacronym{anaclac}{ANACLAC}{Association Nationale des Comités de Langue du Cameroun}

\title{Collaborative construction of lexicographic and parallel datasets for African languages: first assessment}

\author{Elvis MBONING TCHIAZE\\
  NTeALan Research and Development / Makepe, Douala (Cameroun) \\
  \texttt{levismboning@ntealan.org} 
  \\}

\date{}

\begin{document}
\maketitle
\begin{abstract}
Faced with a considerable lack of resources in African languages to carry out work in Natural Language Processing (NLP), Natural Language Understanding (NLU) and artificial intelligence, the research teams of NTeALan association has set itself the objective of building open-source platforms for the collaborative construction of lexicographic data in African languages. In this article, we present our first reports after 2 years of collaborative construction of lexicographic resources useful for African NLP tools.
\end{abstract}

\section{Introduction}
With the exception of a few which are in the process of digitalization, most African languages are very poorly endowed, from a computer point of view, the existing resources are generally in editorial format. The creation of datasets for these languages is the subject of almost no study in \gls{nlp}\footnote{These languages, like most of the subsaharan languages, which are hardly the focus of studies in fields like NLP.} and \gls{ai}, would not only be the basis of an innovative project but above all, subject of great advances. Such project can be an excellent source of data for the various research teams working on these languages. It would indeed be for them and for us as well, an excellent starting point for the realization of various types of research project: development of the \gls{nlp} tools (tokenizer, lemmatizer, stemmizer, tagger, morpho-syntactic analyzer, semantic analyzer, etc.), \gls{nlu} (chatbot, language intention detection), and \gls{ai} (automatic translator, spell checker, sentiment analyzer, document classification, etc.).\\

In this article, we want to share a brief statistical review of our current lexicographic resources from our collaborative platforms. We will give some current limitations of our XND format and the redesign project to consider this year.

\section{Overview of NTeALan project}

Originally initiated as a project in 2017, \gls{ntealan} association initial orientation was to transpose all the latest advances of \gls{nlp}, \gls{nlu} and \gls{ai} to african languages, in terms of operational linguistic tools and resources (useful for African national languages), at the service of research, businesses and populations. Thus, \gls{ntealan}'s \cite{tchiaze_ntealan_2019} aim is to create collaborative and open-source platforms to enable Africans, united in linguistic communities, to be themselves actors in the digitization of their languages and cultures. To achieve this, we decided to rely on the full-fledged collaboration-based model (from Experience Models) proposed by \cite{holtzblatt_7_2017}, because we believe speakers of one language, brought together in a community of contributors, would combine their social, cultural and professional knowledge to promote the technological development and popularization of their languages.\\

Our work, supported by our partners's contribution, has enabled us, in 3 years of existence, to conduct several projects\footnote{NTeALan's publications: \href{https://ntealan.org/publication}{https://ntealan.org/publication}}. They are some of the main tools we developed:

\begin{itemize}
    \item The lexicographic data management platform: made up of an open-source REST API \footnote{\href{https://apis.ntealan.net/ntealan/dictionaries} {https://apis.ntealan.net/ntealan/dictionaries}} and a management web application, it allows us to create, store and distribute/share the platform's linguistic resources. To date, we have more than 10 sub-Saharan languages variously equipped and processed by our systems, including several \gls{nlp} tools.
    \item The collaborative dictionary platform \footnote{\href{https://ntealan.net/dictionaries}{https://ntealan.net/dictionaries}}: used by linguistic communities and members of the association to consult, comment and discuss on the articles of our existing dictionaries\footnote{Article, in the context of lexicography, is considered as the content of language dictionary entry (lemmatized form of word).}. This platform exploits the resources produced thanks to another lexicographic data creation platform \footnote {\href{https://ntealan.net/dictionaries-platform}{https://ntealan.net/dictionaries-platform}}, both are REST API dependent.
    \item Several REST and websocket APIs to make public and distribute \gls{nlp} and \gls{nlu} processing services in African languages. 
\end{itemize}

Most of this work is still functional prototypes that we want to enrich in order to make people outside NTeALan benefit from them.

\section{Context of the work}
\label{sec:length}
Each lexical resource management platform has its own model for structuring and presenting data, this is the case for the following platforms: {\tt Kosh} \cite{kosh2019}, {\tt ELEXIS Dictionary Service}, {\tt Djibiki} \cite{ELX06-024}. The XML format (mainly the TEI and LMF standards) is today a choice of reference for the structuring of linguistic, lexicographic and terminographic data. We can also cite the TEI Lex-0 \cite{romary_tei_2018} and Lexicog (OntoLex Lemon Lexicography from W3C) formats, which are frequently used to codify lexicographic resources. Unfortunately, these standards are often not adapted to represent and describe certain morpho-syntactic peculiarities of African languages \footnote{Certainly for the same reasons, none of these authors have tried these formats for African languages.}. Indeed, several linguistic phenomena, such as the concept of nominal class, the notion of clicks and the management of the translation and the localization of the dialect variants of the entry of an article, are not dealt with explicitly, despite all the needs expressed in the subject \footnote{Note that it is nevertheless possible in these standards to add new formalisms (tags and attributes) in addition to the existing classes. Although this extensibility is possible, their implementation will find reservations in the specification of the linguistic properties and the worldviews concerned.}. \\

Our approach of creating lexical resources is based on the collaboration model. Indeed our current platforms are essentially collaborative. They allow native speakers, experts and teachers of each language to be themselves actors and producers of content for their languages and culture, this was our purpose at the creation of NTeALan's dictionary platforms. However, given the shortcomings observed, we are aware that these platforms should be improved. Thus, within the framework of this project, our approach will take into account several upstream technological factors affecting all our existing platforms, both in terms of functionality, performance and respect of the \gls{rgpd2} rules.\\

To contextualize our work, a first collaborative platform \cite{mboning_ntealan_2020} was set up in 2018. Primary data were extracted from old bilingual lexicographical work available in open access on the internet \cite{mboning_analyse_2016, mboning_linguistique_2018}. In order to facilitate the processing and distribution of our lexicographic data, and based on the state of the art (\cite{prinsloo2012devices,prinsloo2012lexicography,heid2014natural,mangeot_informatisation_2011,el2011referentiel}, etc.) on lexicography in Africa, we created a proprietary structuring model that we called \gls{xnd} \cite{mboning_tchiaze_building_2020}, based on the \gls{xml} format. \\

\begin{figure}[htp!]
    \centering
    \includegraphics[width=220pt]{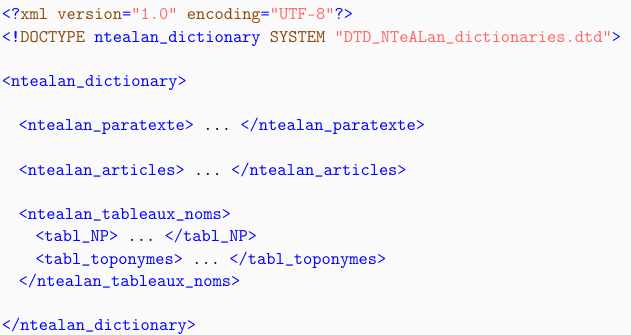}
    \caption{Formal structure of NTeALan's XND format)}
    \label{fig:xnd}
\end{figure}

XND format (cf. figure \ref{fig:xnd}) allowed us to encapsulate and codify many bilingual dictionaries in African languages from different data sources (PDF, Word, XML, Excel files; SIL Toolbox format; HTML pages). These digital dictionaries are currently distributed free of charge on our online REST API \footnote{\href{https://apis.ntealan.net/ntealan/dictionaries}{https://apis.ntealan.net/ntealan/dictionaries}} and are available on the collaborative dictionaries platform \footnote{\href{https://ntealan.net/dictionaries-platform}{https://ntealan.net/dictionaries-platform}}. 

\section{First statistics on current lexicographic data}

After more than 2 years of existence, the data on the platform has grown from 12,000 article entries to over 32,000 articles (equivalent to over 108054 unique words). To be more precise, 18 dictionaries were created with articles between 0 and 12000 . The table \ref{tab:dico_stats}  and the figure \ref{fig:graph_dicos} below give a global and detailed overview of the current statistics \footnote{For the table \ref{tab:dico_stats}: contrib. = user contributions, Dico = dictionaries, consult. = number of articles viewed, comment. = number of comments on the articles.}.

\begin{table}[ht!]
\centering
\begin{tabular}{lcccc}
    \hline \textbf{dico} & \textbf{articles} & \textbf{comment.} & \textbf{consult.} & \textbf{contrib.}  \\ \hline
     18 & 32090 & 99 & 13958 & 156 \\ 
    \end{tabular}
     \caption{General statistics of dictionaries on the NTeALan platform (Record of 05/02/2021)}
    \label{tab:dico_stats}
\end{table}

\begin{figure}[ht!]
    \centering
    \includegraphics[width=230pt]{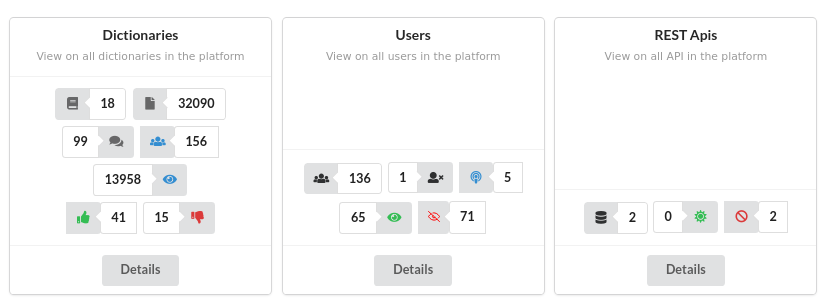}
    \caption{Global view extract from platform (Record of 05/02/2021)}
    \label{fig:global_stats}
\end{figure}

\begin{figure*}[ht!]
    \centering
    \includegraphics[width=410pt]{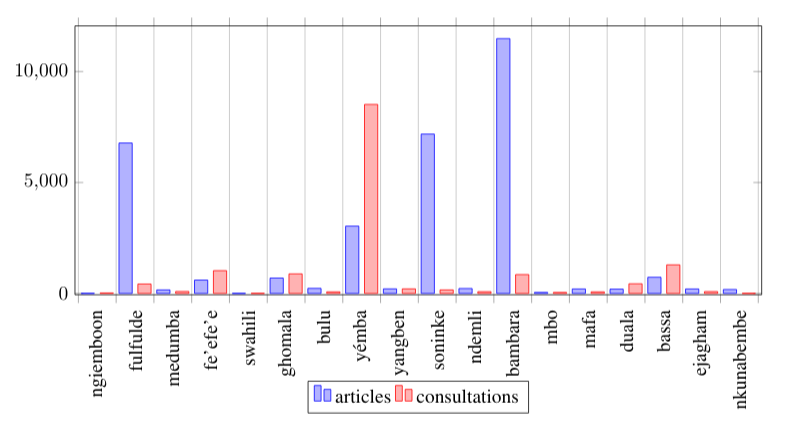}
    \caption{Current frequency of articles in each dictionary in the platform: this distribution shows an average of approximately 1000 articles and ~500 users consultations per dictionary. (Record of 05/02/2021)}
    \label{fig:graph_dicos}
\end{figure*}

\begin{figure*}[ht!]
    \centering
    \includegraphics[width=400pt]{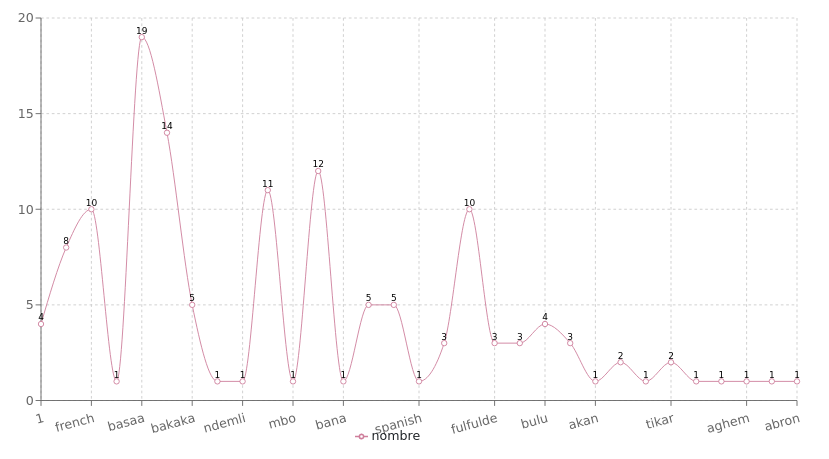}
    \caption{Number of speakers and contributors to the platform grouped by mother tongue (Record of 05/02/2021)}
    \label{fig:graph_user}
\end{figure*}

\begin{figure*}[ht!]
    \centering
    \includegraphics[width=400pt]{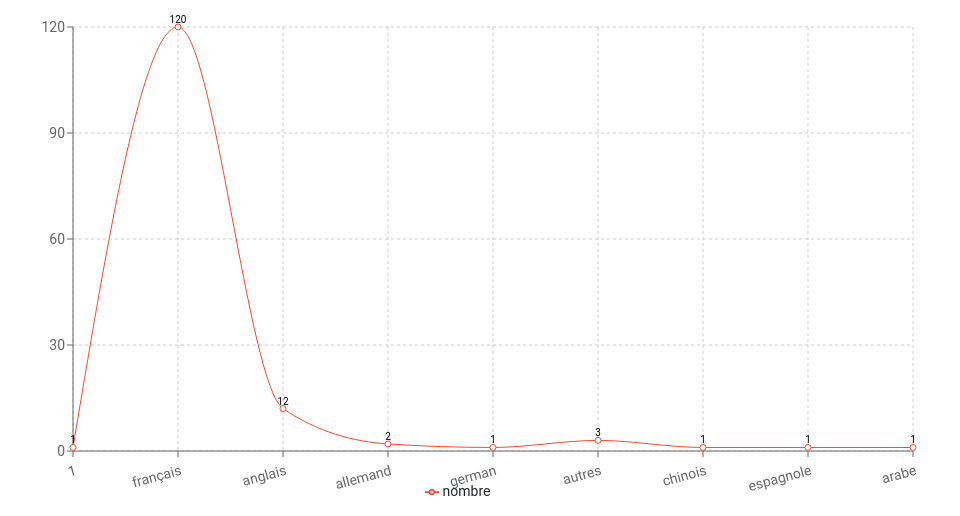}
    \caption{Platform contributors grouped by their official language (Record of 05/02/2021)}
    \label{fig:graph_user_lang}
\end{figure*}

These resources were produced by contributors from several countries. To date, there are more than 135 users / contributors on the platform, with a total of 18 African languages processed and more than 6 foreign languages. At the initiative of the NLP course for Master degree students of the African languages and linguistics (LAL) department of the University of Douala, more than 10 dictionaries have been initialized and are constantly maintained. This student mobilization allowed many of their classmates to join us. Figures \ref{fig:graph_user} and \ref{fig:graph_user_lang} summarize the linguistic diversity of users of the platform. \\

\begin{figure*}[ht!]
    \centering
    \includegraphics[width=400pt]{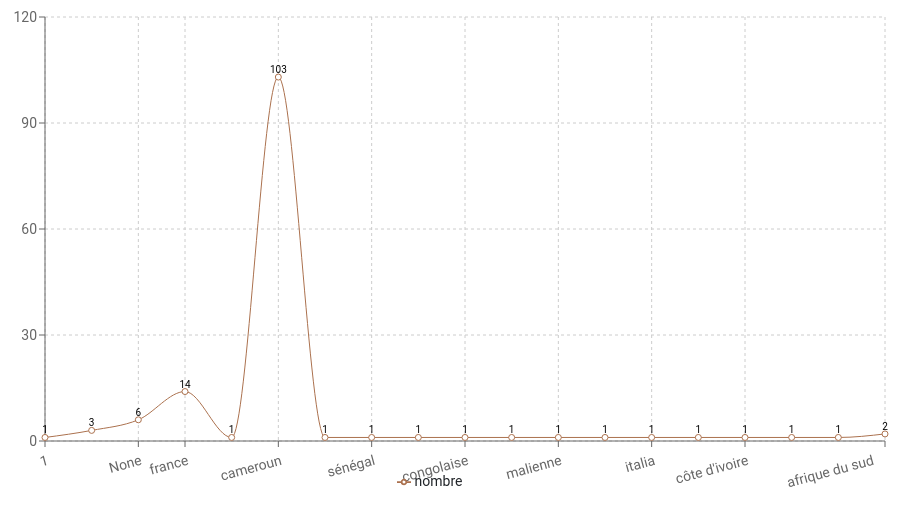}
    \caption{Origins of our contributors to the platform. 'None' represent people who wished to remain anonymous  (Record of 05/02/2021)}
    \label{fig:graph_user_origin}
\end{figure*}

On the other hand, Western languages (spoken as a second or first language in Africa) are fairly well represented (cf. figure \ref{fig:graph_user_origin}). No doubt with good reason, the French language occupies the first place, forwarded by English and German. This can be explained by the fact that these languages are mainly spoken in Central Africa (stronghold of the project).\\

Our platforms indicate that our main contributors are mainly from Cameroon and France. The profiles of these contributors also varied. To date about 43\% of women and 57\% of men are represented in 14 countries in Africa and Europe.

\section{Limitations of the XND format}

The simplicity of the XND format as desired from the start served as exemplary the first waves of dictionary Xmlization, created by the communities of local speakers and students. However, in its current state, this format does not allow us to project on the collaborative model of synchronous or asynchronous resource creation. There are several reasons for this finding:

\begin{itemize}
    \item The linguistic properties of the 4 major families of African languages are under-represented in the current format. Yet these allow them to share, according to their family to which they belong, the same socio-cultural code. Between families, subfamilies, language groups and subgroups, connections of all kinds can be clearly identified. They are full of both linguistic and cultural information that could enrich our different lexicographic content and therefore the current format.
    \item The management of communities of contributors has not been codified in this format, although it is taken into account in the dictionary platform. Metadata relating to collaboration information should at best be an integral part of XML data for better traceability useful to the end user and NLP tools.
    \item The management of external applications and standardization were at the heart of the challenges of defining the XND format. But, faced with the increasing scale of our platforms and its internationalization, the internal structure (tag names and attributes) must be reviewed to better align with existing standards, while keeping its specificity.
    \item The search engine applied to this format is not so effective in finding information desired by the user. Is it an algorithm problem ? It must be said that to date, the implementation of complex cross-search processes has still not been done: the manipulation of XML data in a database remains problematic. \\
\end{itemize}

The XND format must evolve to better serve African languages in their entirety. It must also continue to respect the 3 basic principles {\tt representation, simplification and extensibility} which underpinned its creation. Should we stay on an XML formalism ? improve or enrich it ? or test other modes of knowledge representation ?

\section{The future of the XND format}

The future of XND format will be more semantic and communities based structure following the state of the art. The modeling proposed by \cite{bosch2007towards} turns out to be particularly interesting for us in different aspects: \\

\begin{itemize}
    \item First, the context: the authors were called upon to do pioneering work since they were the first to develop a model for describing lexical entries specific to the Bantu languages of South Africa. We are working in a similar context since there is not yet, to our knowledge, a specific and standard description format for some of the languages described in our dictionaries. Of course, we are not talking about a general model (which already exists) applicable to different languages, but a specific data format that takes into account the particularities of the Mandingue and semi-Bantu languages concerned. etc. 
    
    \item linguistic information: the description method proposed by \cite{bosch2007towards} can take into consideration various types of linguistic information (nominal classes for example) specific to Bantu languages, which are also found in certain languages on which we work: duala, bassa, yemba,  medumba, ghomala, etc. This particularity represents a great advantage since most of the existing models only cover part of the aspects specific to African languages, more precisely those treated in our dictionaries (duala, bassa, yemba, fe'efe'e, medumba, etc.). 
    
    \item the approach: the proposed method is largely in line with the approach we want to implement as part of our project to overhaul the NTeALan dictionary format, in particular a semantic modeling in which the dictionary articles will be structured around the meanings of the terms (the meaning constitutes the entry). Indeed, we wish to develop an internal structuring of the articles of our dictionaries around the meaning, via graphs connected to each other. To do this, like \cite{faass2014general}, we opted for the highlighting of the meaning at the entry of the articles of the dictionaries. As for the generation of graphs, it will be possible thanks to the use of OWL 2 technologies. \\
\end{itemize}

\section{Conclusion}

In this article, we presented the work of the research team of the NTeALan association. More precisely, a statistical assessment of the resources built in collaboration with our communities of contributors. It emerges that the codification format (XND) of our lexicographic resources must evolve to better cover our collaborative community-based model in order to better satisfy our users which is useful for the implementation of NLP tools for African languages. This will be our next challenge.

\section*{Acknowledgments}

This work was made possible thanks to all the voluntary contributors of the NTeALan association and its research team.

\bibliography{eacl2021}
\bibliographystyle{acl_natbib}

\appendix


\printglossaries

\end{document}